\pdfoutput=1

\documentclass[11pt]{article}

\usepackage[final]{acl}

\usepackage{times}
\usepackage{latexsym}

\usepackage[T1]{fontenc}

\usepackage[utf8]{inputenc}

\usepackage{microtype}

\usepackage{inconsolata}

\usepackage{graphicx}
\usepackage{enumitem}
\usepackage{textcomp}
\usepackage{amsmath}
\usepackage{caption}
\usepackage{subcaption}
\usepackage{array}
\usepackage{multirow}
\usepackage{float}

\newcolumntype{C}[1]{>{\centering\let\newline\\\arraybackslash\hspace{0pt}}m{#1}}

\title{A Reproducibility Study on Quantifying Language Similarity:  \\The Impact of Missing Values in the URIEL Knowledge Base}

\author{Hasti Toossi$^\dagger$, Guo Qing Huai$^\dagger$, Jinyu Liu$^\dagger$, Eric Khiu$^*$, \\
\textbf{A. Seza Doğruöz}$^\#$, \textbf{En-Shiun Annie Lee}$^{\dagger,\ddagger}$\\
$^\dagger$ University of Toronto, Canada 
$^*$ University of Michigan, USA \\
$^\#$ LT3, IDLab, Universiteit Gent, Belgium
$^\ddagger$ Ontario Tech University, Canada \\
\texttt{hasti.toossi@mail.utoronto.ca \quad as.dogruoz@ugent.be \quad annie.lee@ontariotechu.ca}
}

\begin{document}
\maketitle
\begin{abstract}
In the pursuit of supporting more languages around the world, tools that characterize properties of languages play a key role in expanding the existing multilingual NLP research. In this study, we focus on a widely used typological knowledge base, URIEL, which aggregates linguistic information into numeric vectors. 
Specifically, we delve into the soundness and reproducibility of the approach taken by URIEL in quantifying language similarity. 
Our analysis reveals URIEL's ambiguity in calculating language distances and in handling missing values. 
Moreover, we find that URIEL does not provide any information about typological features for 31\% of the languages it represents, undermining the reliabilility of the database, particularly on low-resource languages. 
Our literature review suggests URIEL and lang2vec are used in papers on diverse NLP tasks, which motivates us to rigorously verify the database as the effectiveness of these works depends on the reliability of the information the tool provides. 

\end{abstract}

\section{Introduction}

Categorizing and quantifying variations and similarities between languages is critical for applications such as building multilingual large language models \citep{xia2020predicting, team2022NoLL}, examining the effects of cross-lingual transfer \citep{lin2019choosing}, understanding code-switching between languages \citep{dogruoz-etal-2021-survey,dogruoz-sitaram-2022-language}, selecting pivot languages when translating from one language to another \citep{wu2007pivot}, or sharing language tools \citep{strassel2016lorelei}.  However, there is no consensus on how to measure the similarity between languages due to the difficulty and subjectivity involved in assessing various aspects of languages. This challenge becomes even more pronounced when dealing with low-resource languages \citep{joshi2020state}, where limited linguistic knowledge is available to researchers.

\begin{figure}[!ht]
\begin{center}
\includegraphics[width=0.45\textwidth]{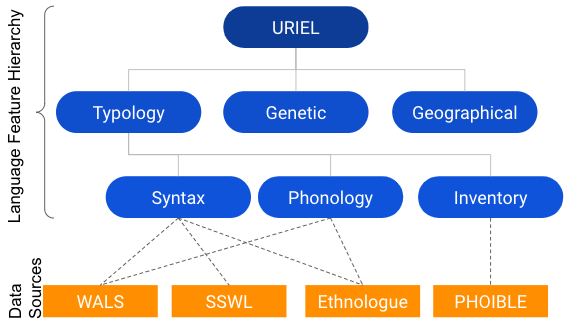} 
\caption{URIEL Feature Hierarchy and Data Sources.}
\label{fig.uriel}
\end{center}
\end{figure}

URIEL is a knowledge base that aggregates linguistic information for 4,005 languages from various data sources (Figure~\ref{fig.uriel}) and computes distances based on this information. The lang2vec tool provides an interface for querying URIEL \citep{littell2017uriel}. In many of the 198 citations of URIEL and lang2vec, the distance values and feature vectors provided by URIEL have been used to quantify language similarity and categorize language features.

In this study, we analyze the URIEL database to assess its capacity as a resource for quantifying language similarity. We evaluate the reproducibility and validity of the methodology employed in calculating language similarity measurements. We also examine the language and feature coverage of URIEL, which affects the meaningfulness of the vectors and distance values.

In addition, we conducted a literature review of papers that cite URIEL to gain a better understanding of the influence of URIEL in these works. 

\section{Methodology for Reproducibility}

\subsection{Description of URIEL}
For a pair of languages, URIEL computes the distance of the corresponding features through the following steps:  
\begin{enumerate}
    \item Collect information from various sources for a specific feature.
    \item Take an aggregate of the different sources for a single feature.
    \item Compute the distance of the feature vectors of the two languages.
\end{enumerate}

\paragraph{URIEL knowledge base} URIEL unifies information from various sources (Figure~\ref{fig.uriel}), such as WALS \citep{wals}, SSWL \citep{koopman2009syntactic}, PHOIBLE \citep{phoible}, Ethnologue \citep{ethnologue}, and Glottolog \citep{hammarstrom2021glottolog}.  The \emph{features} of a language are broken down into three types:  
\begin{enumerate}
    \item Typological features \texttt{syntax}, \texttt{phonology} and \texttt{inventory}, which describe the corresponding linguistic characteristics of the language.
    \item Phylogenetic feature \texttt{family}, which specifies the language families to which the language belongs.
    \item Geographical feature \texttt{geography} for the approximate location where the language is most commonly spoken in the world.
\end{enumerate}

All {features} are described using binary (0 or 1) vectors to represent language facts. Missing values are marked by ``\texttt{-{}-}''.  

For each feature, different vectors are provided depending on the source. For instance, URIEL provides \texttt{syntax} vectors sourced from each of WALS, SSWL, and Ethnologue. Similarly, other feature vectors are derived from multiple sources.

\paragraph{Aggregating sources}

Since the information for each feature can be taken from several sources, URIEL uses three aggregation methods to consolidate feature information: union, average, or $k$-nearest neighbours ($k$NN).  

For the union aggregation, denoted using the union operator ``\texttt{|}'', each feature is set to 1 if any of the sources for that feature has a value of 1. If the feature value is 0 in all sources, the feature is set to 0. If the feature value is missing in all sources, the union result has a missing entry, denoted by ``\texttt{-{}-}''.

For the average aggregation, each entry is the mean across all sources in which it appears. This result is a value between 0 and 1, with a non-binary value if there are disagreements among the sources. The feature is missing, denoted ``\texttt{-{}-}'', if the value is missing in all sources. 

Lastly, for the $k$NN aggregation, the missing values are predicted based on languages similar in terms of genetic, geographic and featural distances. It is unclear how aggregation is done for $k$NN, as the details are omitted from the URIEL paper. \citet{littell2017uriel} writes: ``We will describe these procedures, the exact notions of distance involved, alternative prediction methods that we also investigated, and their results in more detail in a future article.''

\paragraph{Computing language distances}
For each language pair, URIEL provides pre-calculated distance values based on the aggregated feature vectors. 
While the exact methodology for distance calculations is not specified in the URIEL paper \citep{littell2017uriel}, additional documentation for URIEL and lang2vec provides two different distance calculation methods. 

The lang2vec documentation\footnote{lang2vec is the Python tool developed by the authors for querying URIEL. \href{https://github.com/antonisa/lang2vec}{https://github.com/antonisa/lang2vec}} uses cosine distance to compute distances between feature vectors. The cosine distance $D_C$ between two vectors $u$ and $v$ is defined as
\begin{align}
D_C(u, v) := 1 - S_C(u, v)
\end{align}
where $S_C$ is cosine similarity defined by
\begin{align}
S_C(u, v) := \frac{u \cdot v}{\lVert u \rVert\lVert v \rVert}. \label{eq:sim}
\end{align}

On the other hand, the URIEL documentation\footnote{\href{http://www.cs.cmu.edu/~dmortens/projects/7_project/}{http://www.cs.cmu.edu/\texttildelow dmortens/projects/7\_project/}} defines a distance equivalent to angular distance. The angular distance $D_\theta$ between two vectors $u$ and $v$ is defined as
\begin{align}
D_\theta(u, v) := \frac{1}{\pi}\arccos(S_C(u, v))
\end{align}
where $S_C$ is the same cosine similarity defined in (\ref{eq:sim}).

Note that the value of $D_\theta(u, v)$ can range between $0$ and $0.5$ because all feature values are positive. However, distances in URIEL range between 0 and 1, with ``0 representing identity and 1 being as far apart as two languages can be'' based on the URIEL documentation. Therefore, it is reasonable to assume that this distance metric is regularized to $2D_\theta(u, v)$. 


\begin{table*}[!ht]
\begin{centering}
\resizebox{\textwidth}{!}{%
\begin{tabular}{||c c|c c c|c c c||}
\hline
\multirow{2}{*}{\begin{tabular}{c}Aggregate \\ Vector\end{tabular}} & \multirow{2}{*}{\begin{tabular}{c}Distance \\ Metric \end{tabular}} & \multicolumn{3}{c|}{All Languages} & \multicolumn{3}{c||}{Languages with Non-Empty Feature Vectors} \\
& & \texttt{syntactic} & \texttt{phonological} & \texttt{inventory} & \texttt{syntactic} & \texttt{phonological} & \texttt{inventory}\\
\hline
\hline
\texttt{union} & {cosine} & 23.90\% & 61.62\% & 40.04\% & 14.24\% & 34.28\% & 0.07\%\\
\texttt{union} & {angular} & \textbf{93.36\%} & \textbf{95.42\%} & \textbf{99.45\%} & \textbf{95.89\%} & \textbf{87.76\%} & \textbf{98.52\%}\\
\texttt{average} & {cosine} & 23.95\% & 61.62\% & 40.04\% & 14.38\% & 34.28\% & 0.07\%\\
\texttt{average} & {angular} & 89.82\% & 95.21\% & 90.53\% & 88.92\% & 86.90\% & 71.23\%\\
\texttt{knn} & {cosine} & 0.39\% & 1.45\% & 0.12\% & 0.39\% & 1.45\% & 0.12\%\\
\texttt{knn} & {angular} & 2.46\% & 2.53\% & 9.70\% & 2.46\% & 2.53\% & 9.70\%\\
\hline
\end{tabular}
}
\caption{Percentage of all language pairs with reproducible distances (up to 2 decimal places) using each method.}
\label{tab.reproduce}
\end{centering}
\end{table*}

\section{Results}
\label{sec:results}

\subsection{Reproducibility Study}\label{sec:repro}
We attempted to reproduce the pre-calculated distance provided by URIEL for each language pair. This involved reproducing the aggregation step and the distance calculation step using the feature vectors provided by URIEL.  We used the aggregated feature vectors from URIEL as the basis for the distance computations.

\paragraph{Reproducing aggregated vectors}

While we successfully reproduced the first two aggregation vectors (union and average), we were unable to replicate the exact $k$NN aggregation vector because the necessary details were not provided.  

\paragraph{Reproducing distance calculations}
As mentioned earlier, URIEL provides pre-computed distances for each language pair based on different feature vectors (particularly \texttt{syntactic}, \texttt{phonological}, and \texttt{inventory}). However, the methodology used to calculate these distances is unclear in the documentation. We aimed to reproduce these provided distance values to infer the methodology used.

There are three ambiguities in the documentation regarding distance computations:
\begin{enumerate}
    \item Which aggregated vector is used; \texttt{union}, \texttt{average}, or \texttt{knn}?
    \item Which distance metric is used; cosine distance $D_C(u, v)$ or regularized angular distance $2D_\theta(u, v)$?
    \item How are the missing feature values treated?
\end{enumerate}
We found that among possible methods of treating missed values, the following method aligns with the pre-computed distances closely:
\begin{itemize}
    \item If every value in a feature vector is missing, replace it with a vector of the same length containing only 1's.
    \item If some, but not all, values in a vector are missing, replace the missing values with 0.
\end{itemize}

Using this method for treating missing values, we calculated the distances for each language pair using all possible combinations of aggregation methods and distance metrics. 

The percentage of all language pairs whose distance can be reproduced with each set of choices is shown in Table \ref{tab.reproduce} (``All Languages'' section)\footnote{URIEL provides \texttt{phonological} and \texttt{inventory} distances up to 4 decimal points. However, reproducibility suffers when using more than 2 decimal points.}. The highest percentage of reproducible distances was achieved using regularized angular distance with \texttt{union} vectors.

A similarly high percentage of distances could be reproduced by using regularized angular distance with \texttt{average} vectors instead of \texttt{union} vectors. This can be explained by noting that the \texttt{union} and \texttt{average} vectors are identical for many languages. Corresponding \texttt{average} and \texttt{union} vectors are equal when all available sources agree on the relevant features of a language. Specifically, \texttt{syntax\_union} and \texttt{syntax\_average} are equal for $95.23\%$ of languages, \texttt{phonology\_union} and \texttt{phonology\_average} for $99.73\%$ of languages, and \texttt{inventory\_union} and \texttt{inventory\_average} for $91.59\%$ of languages.

Additionally, in Table \ref{tab.reproduce} (``Languages with Non-Empty Feature Vectors'' section), we consider the reproducibility of distances for language pairs where both languages have non-empty feature vectors. This is relevant because all empty vectors are considered identical for distance purposes.

We conclude that regularized angular distance with \texttt{union} vectors is the most likely method used to calculate the pre-computed distance vectors provided by URIEL. However, some distance values could not be  reproduced using this or any other method we tried. There are no clear factors causing the irreproduciblity of certain distance values.

\subsection{Analysis of Feature Coverage}\label{sec.coverage}

URIEL provides feature vectors for 4,005 languages, as well as corresponding distance values for all pairs of these languages (16,040,025 pairs). 

However, a large number of features in these vectors have missing values, and many feature vectors are completely empty, indicating that every feature in these vectors is missing. This raises concerns about the meaningfulness of the distance values provided for such languages, as the vectors contain no information to distinguish these languages.

Out of the 4,005 languages, 1,735 (43.32\%) have empty \texttt{syntax\_union} vectors, 2,914 (72.76\%) have empty \texttt{phonology\_union} vectors, and 2,534 (63.27\%) have empty \texttt{inventory\_union} vectors. Furthermore, 1,251 (31.24\%) languages have no feature information at all, meaning they have empty vectors for \texttt{syntax\_union}, \texttt{phonology\_union}, and \texttt{inventory\_union}. Some languages have empty vectors in one or more of these three categories, but not all. 
\begin{figure}[!t]
\centering
    \begin{subfigure}[b]{0.5\textwidth}
        \includegraphics[width=\textwidth]{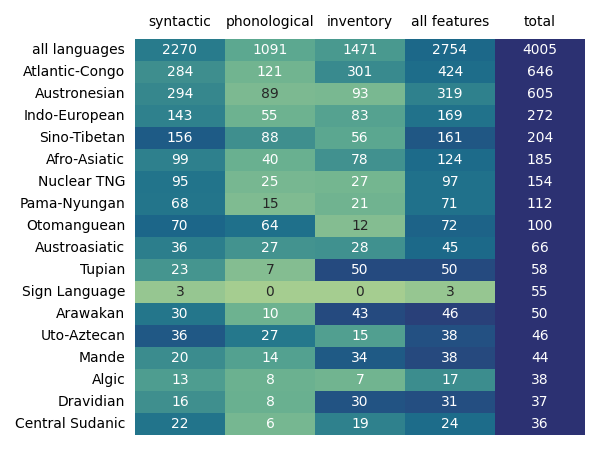}
        \caption{In the 20 largest language families.}
        \label{tab.coverage}
    \end{subfigure}
        \begin{subfigure}[b]{0.5\textwidth}
        \includegraphics[width=\textwidth]{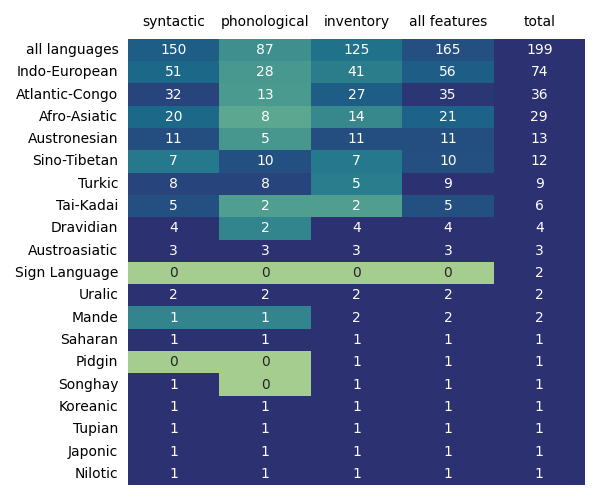}
        \caption{In the top 200 most spoken languages.}
        \label{tab.coverage.appendix}
    \end{subfigure}
    \caption{Number of languages with non-empty \texttt{union} feature vectors}
\end{figure}

Figure \ref{tab.coverage} shows the number of languages with non-empty \texttt{union} feature vectors in each of the 20 largest language families. The column labelled ``all features'' represents the number of languages with non-empty \texttt{union} feature vectors in at least one of the categories. The ``total'' column shows the total number of languages from each language family included in URIEL. The shading indicates the percentage of languages with non-empty vectors compared to the total number of languages in the corresponding language family.   

Similarly, Figure \ref{tab.coverage.appendix} focuses on the top 200 most spoken languages in the world\footnote{Excluding Bajjika, the 103rd most spoken language, which is missing from URIEL.}, as identified by Ethnologue 2023. Figure \ref{tab.coverage.full} presents this information for all language families in URIEL.

\subsection{Distribution of Non-Missing Features} 

\begin{figure}[!t]
\centering
    \begin{subfigure}[b]{0.5\textwidth}
        \includegraphics[width=\textwidth]{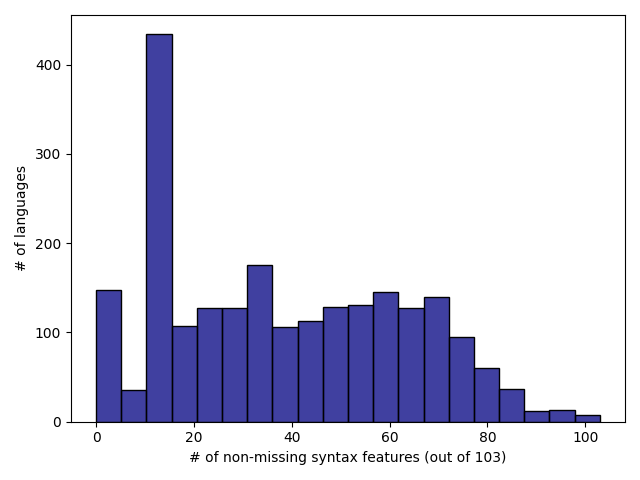}
        \caption{non-missing features in \texttt{syntax\_union}}
        \label{tab.non-missing.syntax}
    \end{subfigure}
    \begin{subfigure}[b]{0.5\textwidth}
        \includegraphics[width=\textwidth]{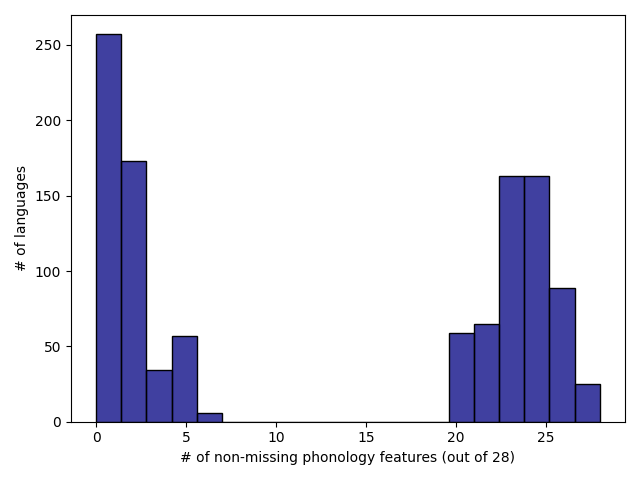}
        \caption{non-missing features in \texttt{phonology\_union}}
        \label{tab.non-missing.phonology}
    \end{subfigure}
    \begin{subfigure}[b]{0.5\textwidth}
        \includegraphics[width=\textwidth]{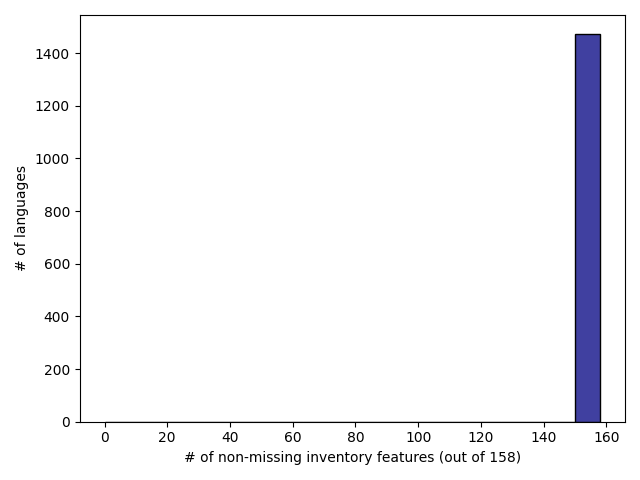}
        \caption{non-missing features in \texttt{inventory\_union}}
        \label{tab.non-missing.inventory}
    \end{subfigure}
    \caption{Distribution of languages based on the number of non-missing features in the \texttt{union} vector for each category, excluding languages with empty feature vectors.}
    \label{tab.non-missing}
\end{figure}

In section \ref{sec.coverage}, we discussed languages with empty feature vectors, i.e., languages that lack any feature information in a given category. We found that these languages constitute a large portion of all languages in the URIEL dataset.

To better understand the feature coverage of the remaining languages, we will now exclude the languages with empty feature vectors. In Figure \ref{tab.non-missing}, we visualize the distribution of the remaining languages based on the number of feature values provided in their \texttt{union} vectors for each category. 

In Figure \ref{tab.non-missing.inventory}, we observe that if any language has a non-empty \texttt{inventory\_union} vector, then this vector contains no missing values. By referencing the original source\footnote{In this case, the relevant source is PHOIBLE, a repository of cross-linguistic phonological inventory data.}, we find that this source provides complete International Phonetic Alphabet (IPA) charts for all languages it covers. Since \texttt{inventory} vectors represent the information from the IPA chart of each language, they do not have any missing values when a complete IPA chart is available.

As depicted in Figure \ref{tab.non-missing.phonology},languages with non-empty \texttt{phonology\_union} vectors generally have either at least 20 or at most 7 phonology features, with no values in between. On the other hand, syntax features exhibit a more even distribution (Figure \ref{tab.non-missing.syntax})  with a peak in the number of languages with 11 to 15 syntax features.

\section{Literature Review}\label{sec.litrev}
\subsection{Methodology of Literature Review}
A structured search strategy was implemented to gather articles containing citations of URIEL/lang2vec from Google Scholar, sorted by relevance. We then reviewed each paper through a particular process.
First, we read the abstract and introduction of the paper to fill in the summary section, and identified relevant keywords from each paper. Then, we used the search function to find occurrences of “Littel”, “URIEL”, and “lang2vec” to locate where and how URIEL was used in the paper. Finally, we searched for keywords such as “database”, “language distance”, and “WALS” to identify other methods co-existing with or compared to URIEL in the paper.

Following the initial search, duplicated instances of URIEL usage and articles with similar topics were categorized. Further analysis focused on the most cited articles, as well as articles relevant to performance prediction, language distance, and typological feature comparison. Selected articles underwent a full-text review, during which a detailed examination of methodologies, findings and limitations was conducted. 


\subsection{Findings from Literature Review}
Our literature review consists of a comprehensive analysis of 198 citations of the URIEL database up to February 2024. The cited literature focuses on a range of topics, including cross-lingual modelling, performance prediction, and other NLP applications such as document image classification, text-to-speech, and speech recognition 
\cite{adams2019massively,raj2023masr}.

\citet{lauscher2020zero} explored the efficacy of URIEL in cross-lingual modelling, cross-lingual learning, and zero-shot transfer scenarios. \citet{patankar-etal-2022-train}, \citet{xia2020predicting}, and  \citet{srinivasan2021predicting} delved into methodologies for predicting the performance of multilingual NLP models across diverse tasks; these tasks include Natural Language Inference, Machine Comprehension, Named Entity Recognition, Entity Linking, Morphological Analysis, Universal Dependency Parsing, Part Of Speech Tagging, Dependency Parsing, and Machine Translation.

URIEL and lang2vec are also used select the source language in cross-lingual transfer tasks and language translation tasks. 
\citet{lin-etal-2019-choosing} attempt to solve the task of automatically selecting optimal transfer languages as a ranking problem and build models that consider URIEL's language features to perform this prediction. \citet{huang2021improving} use lang2vec to verify model outcomes, evaluate the effectiveness of models across different languages, and analyze the correlation between model outcomes and language distance between the source and target languages in language translation tasks. Aside from language distance computation, \citet{ustun-etal-2020-udapter} integrated lang2vec into models such as BERT and multilayer perceptrons, enhancing their performance across various linguistic tasks.

\citet{adilazuarda2024lingualchemy} demonstrated a way to align lang2vec feature vectors and Multilingual BERT (mBERT) embeddings to explore whether multilingual language models (MLMs), such as mBERT, capture the linguistic constraints defined by URIEL vectors. 
Upon observing that mBERT embeddings and lang2vec vectors strongly correlate, a new method(LINGUALCHEMY) was introduced to align model representations with linguistic knowledge from URIEL vectors. This is achieved by adding an additional URIEL loss term to the regular classification loss. URIEL loss is defined as the mean squared error (MSE) between projected model output and the corresponding URIEL vectors.

Notably, \citet{10.1162/coli_a_00357} highlighted the issue of predicted World Atlas of Language Structures (WALS) values from URIEL exhibiting noticeable clusters, due to biases introduced by family-based prediction of missing values in URIEL.

\section{Conclusion}


In conclusion, in our attempt to reproduce URIEL's ``language distances'', we identified several areas for improvement:

\begin{itemize}
    \item \textbf{Unclear definitions:} The documentation for the definition of distance values provided by URIEL is unclear. Through our attempts, we identified the likely definitions used, but there are some distance values that remain irreproducible for unknown reasons.

    \item\textbf{Missing Values:} When computing distances, missing values in the feature vectors are handled by replacement with $0$. There is no clear justification for this approach, which affects distance values for languages with many missing values (e.g., with a majority being the low-resource languages).

    \item\textbf{Low Coverage:} We found that $31.24\%$ of the languages in URIEL have no linguistic feature information. While language distance values are provided for these languages by URIEL, they are not meaningful due to the empty feature vectors.  While the low coverage leads to a broader issue for low-resource languages, which is difficult to solve, URIEL can address this by providing a \texttt{nan} value in these cases, which would make it clearer to the user when language distance values cannot be meaningfully derived.
\end{itemize}

As demonstrated in our literature review, there are broad use cases for measuring language similarity. By understanding and addressing areas of improvement for URIEL and lang2vec, we can contribute to the progress of research in multilingualism and language diversity, especially for low-resource languages that are not properly represented by these knowledge bases and tools.


\subsection{Future Work}

For future research, we are planning to establish clear guidelines for acceptable levels of missing data in linguistic datasets. Secondly, we aim to refine URIEL specifically for medium- and high-resource languages. For low-resource languages, we will explore alternative similarity measurements, such as conceptual distance or other overlooked linguistic features. Our objective is to advance
computational linguistics research by tackling missing data challenges and improving method applicability across diverse linguistic contexts.





\subsection{Limitations}
The limitation of this research is its reliance on the accuracy and completeness of the URIEL knowledge base when extracting data from its sources. Any inaccuracies or omissions within the URIEL dataset could impact the reproducibility and reliability of our findings.  We did not verify the reliability of the external data sources, nor did we compare them against URIEL. 



\section*{Acknowledgement}
We thank Pratik Nadipelli for his valuable assistance in the literature review, Aditya Khan, Phuong Hanh Hoang, and Mason Shipton for their meticulous efforts in proofreading the paper. Lastly, we appreciate our NAACL24 mentor William Held for his valuable feedback on our first draft. 

\bibliography{langdist}

\appendix

\section{Top 200 Most Spoken Languages}

\begin{figure}[!t]
\centering
    \begin{subfigure}[b]{0.5\textwidth}
        \includegraphics[width=\textwidth]{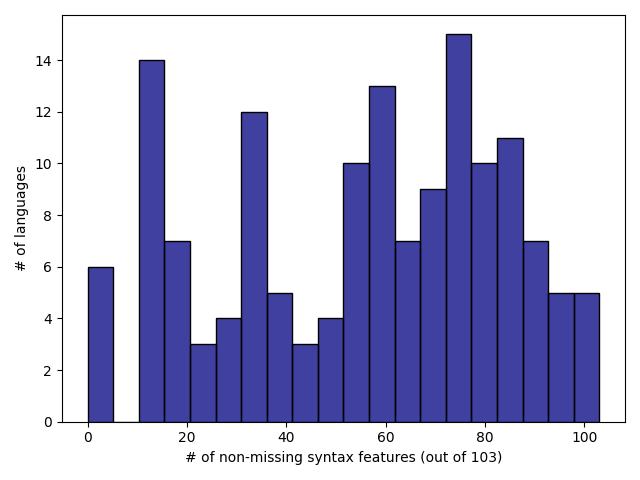}
        \caption{non-missing features in \texttt{syntax\_union}}
        \label{tab.non-missing.syntax.appendix}
    \end{subfigure}
    \begin{subfigure}[b]{0.5\textwidth}
        \includegraphics[width=\textwidth]{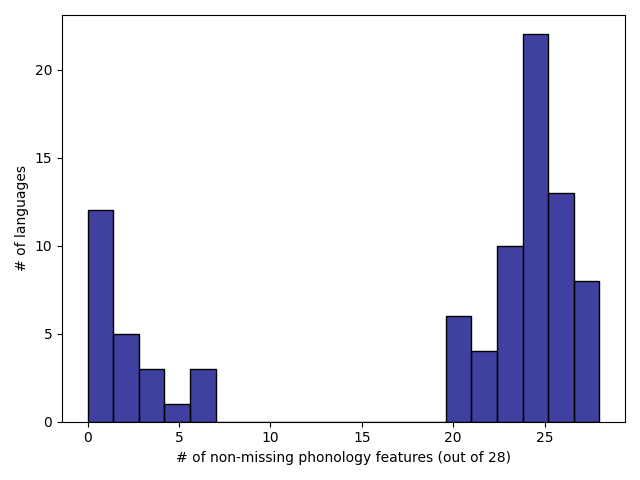}
        \caption{non-missing features in \texttt{phonology\_union}}
        \label{tab.non-missing.phonology.appendix}
    \end{subfigure}
    \begin{subfigure}[b]{0.5\textwidth}
        \includegraphics[width=\textwidth]{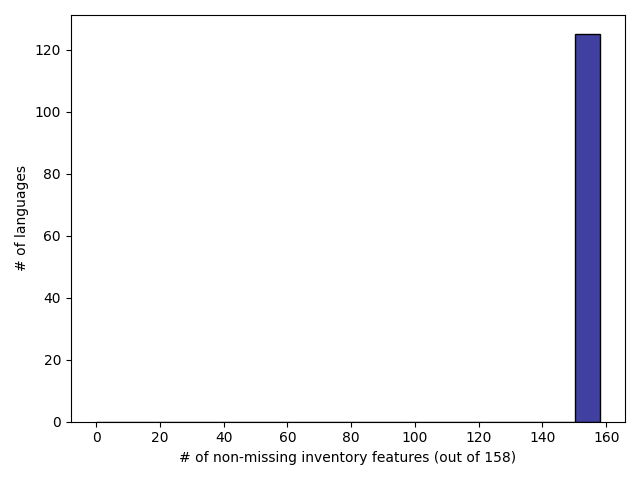}
        \caption{non-missing features in \texttt{inventory\_union}}
        \label{tab.non-missing.inventory.appendix}
    \end{subfigure}
    \caption{Distribution of the top 200 most spoken languages based on the number of non-missing features in the \texttt{union} vector for each category, excluding languages with empty feature vectors.}
    \label{tab.non-missing.appendix}
\end{figure}

Figure \ref{tab.non-missing.appendix} provides information similar to Figure \ref{tab.non-missing} in the main text, but focuses on the top 200 most spoken languages in the world, as identified by Ethnologue 2023, instead of all 4,005 languages in URIEL.  

Note that Bajjika, the 103rd most spoken language in the world (with 12.3M speakers), is missing from URIEL. Consequently, figures \ref{tab.coverage.appendix} and \ref{tab.non-missing.appendix}  include data only for the other 199 languages.


\section{Full Table of Coverage Based on Language Family}
\begin{figure*}[!tp]
\centering
    \begin{subfigure}[b]{0.49\textwidth}
        \includegraphics[width=\textwidth, height=0.95\textheight, keepaspectratio]{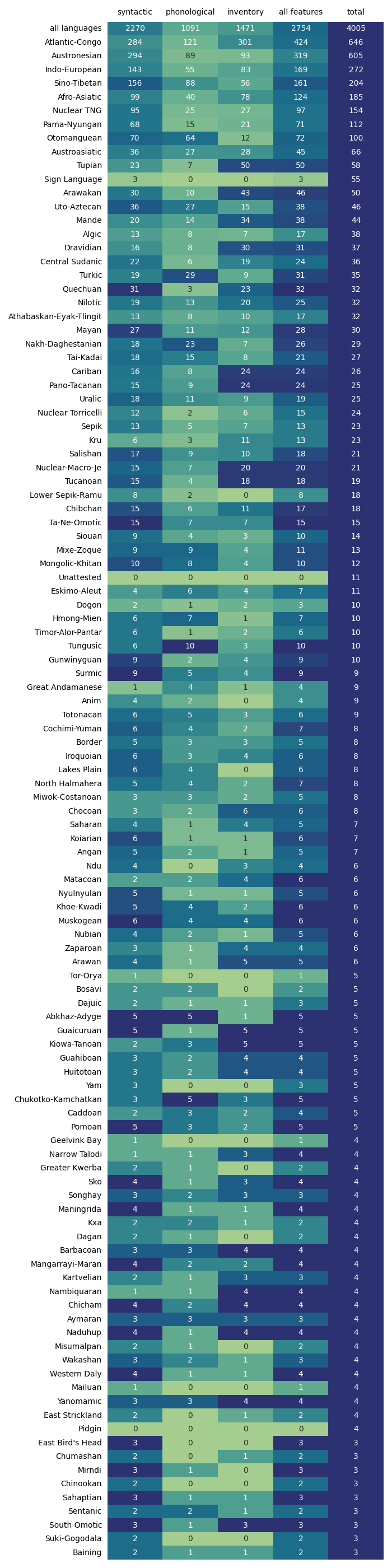}
    \end{subfigure}
    \begin{subfigure}[b]{0.49\textwidth}
        \includegraphics[width=\textwidth, height=0.95\textheight, keepaspectratio]{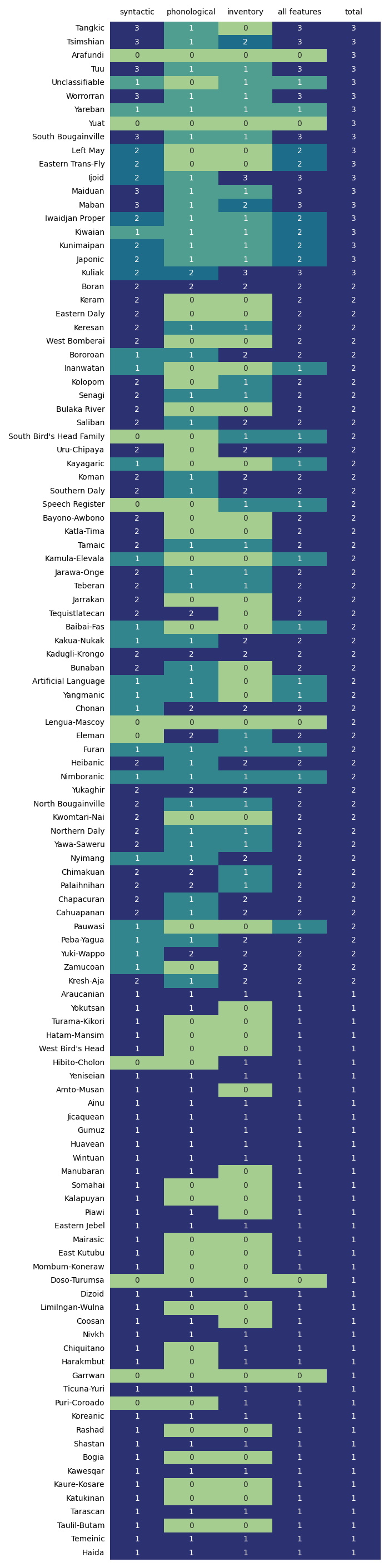}
    \end{subfigure}
    \caption{Number of languages with non-empty \texttt{union} feature vectors in all language families.}
    \label{tab.coverage.full}
\end{figure*}

Figure \ref{tab.coverage.full} shows the number of languages with non-empty feature vectors for each language family in URIEL. Figure \ref{tab.coverage} in the main text is an abridged version that displays only the $200$ largest language families.

\end{document}